\documentclass[11pt]{article}
\usepackage[preprint]{acl}

\usepackage{times}
\usepackage{latexsym}

\usepackage[T1]{fontenc}

\usepackage[utf8]{inputenc}

\usepackage{microtype}

\usepackage{inconsolata}

\usepackage{graphicx}

\usepackage{booktabs}
\usepackage{colortbl}

\usepackage{tabularx}
\usepackage{multirow}
\usepackage{subcaption}
\usepackage{framed}
\usepackage{soul}
\usepackage{amsmath}

\definecolor{grey}{gray}{0.9}

\newcommand{\Hrule}[3][.]{%
  \par\addvspace{#2}%
  \begingroup\color{#1}%
  \hrule
  \endgroup
  \addvspace{#3}%
}

\newenvironment{cframed}[1][gray!40]
  {\def\FrameCommand{\fboxsep=\FrameSep\fcolorbox{#1}{white}}%
    \MakeFramed {\advance\hsize-\width \FrameRestore}}
  {\endMakeFramed}

\newcommand\nl[1]{\textit{#1}}

\definecolor{forestgreen}{rgb}{0.13, 0.55, 0.13}

\title{\texttt{GRUFF}: LLM Pronoun Fidelity, Reasoning, and Biases in German}

\author{
 \textbf{Fabian Mewes\textsuperscript{1,2}} \quad\quad
 \textbf{Anne Lauscher\textsuperscript{2}} \quad\quad
 \textbf{Vagrant Gautam\textsuperscript{3}}
\\
 \textsuperscript{1}JobMatchMe GmbH, Germany \\
 \textsuperscript{2}Trustworthy AI Lab, University of Hamburg, Germany \\
 \textsuperscript{3}Heidelberg Institute for Theoretical Studies, Germany
\\
 \small{
   \texttt{fabmew2000@gmail.com},
   \texttt{anne.lauscher@uni-hamburg.de},
   \texttt{vagrant.gautam@h-its.org}
 }
}

\begin{document}
\maketitle
\begin{abstract}
Third-person singular pronouns have long been used to study stereotypical biases in language models and to test their abilities to reason about reference.
More recently, the interplay between reasoning and bias has been investigated with the task of pronoun fidelity, which assesses models' abilities to correctly reuse a previously-specified pronoun for a discourse entity, independent of other potentially distracting discourse entities mentioned in between.
However, such research focuses on English, a language with limited grammatical gender and almost no gender agreement.
In this paper we contribute a novel, large-scale dataset, \texttt{GRUFF}, to measure pronoun fidelity in German, covering four different gender agreement systems in nouns, and four sets of pronouns.
With this dataset, we show that LLMs display strong grammatical agreement for masculine and feminine entities in the absence of explicit context, but not for neopronouns \textit{xier} and \textit{en}.
In the presence of distractors, however, models degrade on all pronoun sets.
Finally, we show that occupational stereotypes in this context are poorly correlated across grammatical cases and most models, except ones with closely related architectures.
\end{abstract}

\section{Introduction}
\begin{figure}[th!]
    \begin{cframed}
        \noindent\textbf{Grammatical Agreement} (\S\ref{sec:agreement})\\
        \noindent\nl{\textbf{Die Tierärztin} benötigte Kaffee, weil \textbf{sie} sehr früh aufgestanden war. \textbf{Die Tierärztin} sagte, dass \_\_\_ an Wochenenden nicht arbeitet.} \\
        \noindent\nl{(\textbf{The vet} needed coffee as \textbf{she} woke up early. The vet said \_\_\_ did not work weekends.)}
        
        \vspace{0.05in}
        \Hrule[gray!40]{3pt}{5pt}
        \vspace{0.05in}

        \noindent\textbf{Robustness to Distractors} (\S\ref{sec:distractors})\\
        \noindent\nl{Die Tierärztin benötigte Kaffee, weil sie sehr früh aufgestanden war. \textbf{Dier Besitzer*in} war voll, weil \textbf{xier} gerade eine große Mahlzeit gegessen hatte. Die Tierärztin sagte, dass \_\_\_ an Wochenenden nicht arbeitet.} \\
        \noindent\nl{(The vet needed coffee as she woke up early. \textbf{The owner} was full because \textbf{xe} ate a big meal. The vet said \_\_\_ did not work weekends.)}

        \vspace{0.05in}
        \Hrule[gray!40]{3pt}{5pt}
        \vspace{0.05in}

        \noindent\textbf{Biases and Stereotypes} (\S\ref{sec:bias})\\
        \noindent\nl{Die Tierärztin war gesättigt, weil ihr das Essen gut geschmeckt hatte. Die Tierärztin sagte, dass \_\textbf{\textcolor{red}{er}}\_ an Wochenenden nicht arbeitet.}
        
        \noindent\nl{(The vet was full as she had enjoyed the food. The vet said that \_\textbf{\textcolor{red}{he}}\_ did not work weekends.)}

    \end{cframed}
    \caption{Examples with English translations from our proposed \texttt{GRUFF} dataset (\S\ref{sec:gruff}) where the blank should always be filled with \textbf{sie} \textit{(she)}. Different splits of our dataset let us quantify the role of grammatical agreement and robustness to distractors in German pronoun fidelity; incorrect LLM responses reveal stereotypical biases.}
    \label{fig:one}
\end{figure}

Third-person singular personal pronouns (e.g., \emph{he}, \emph{she}, \emph{they} in English) are among the smallest units of language, yet they are important ways to index and construct identity and gender~\cite{lauscher-etal-2022-welcome}. 
Thus, as evidenced by psychological research~\cite[e.g.,][]{mclemore2018minority}, failing to use an individual's pronouns correctly (i.e., \emph{misgendering}) can result in significant psychological harm, and pronouns are a linguistic site where minority stress occurs.
As large language models (LLMs) become increasingly integrated into our daily lives, their ability to navigate these nuances is no longer just a technical requirement, but an ethical imperative.

Researchers in natural language processing (NLP) have started to address this issue by creating resources to evaluate and improve how LLMs handle third-person pronouns~\citep[e.g.,][]{lauscher-etal-2023-em, ovalle-etal-2024-tokenization,hossain-etal-2024-misgendermender}.
One such work, \citet{gautam-et-al-2024-RUFF}, introduced the task of \textit{pronoun fidelity} -- assessing whether, given a context introducing a co-referring entity and pronoun, models are able to reuse the correct pronoun later.
To this end, the authors propose \texttt{RUFF}, a dataset to test model robustness to non-adversarial distractor sentences discussing other entities in between.
Unfortunately, the vast majority of existing research on safe pronoun usage (\emph{i}) focuses exclusively on English~\citep[e.g.,][]{hossain-etal-2023-misgendered, gautam-et-al-2024-RUFF}, or (\emph{ii}) is restricted to testing pronominal bias in machine translation \citep[e.g.,][]{lauscher-etal-2023-em}, despite the reality that LLMs are used by millions of people worldwide in their native languages~\cite{wang2023all}.

In this work, we acknowledge that gendered language also impacts German speakers \citep{huck-2021-thesis} and study pronoun fidelity in German. Here, grammatical gender is marked not just on pronouns, but also nouns and articles, providing stronger cues for gender agreement.
We introduce \texttt{GRUFF}, the first resource for studying pronoun fidelity in LLMs for German, covering four noun agreement systems, and four pronoun sets across four grammatical cases in German---including gender-inclusive noun variants such as the Sternchen and neopronouns like \textit{xier}.
Using \texttt{GRUFF}, we  conduct a comprehensive empirical analysis of LLMs' abilities to correctly track and reproduce pronouns.

We find that LLMs model almost-perfect gender agreement for traditional pronouns in German, independent of explicit context specifying the pronoun.
However, LLMs only show agreement with neopronouns when they are explicitly specified.
As in English, discourse complexity in the form of distractors dramatically affects performance, despite the additional gender agreement.
Gender marking likely also mitigates the effects of occupational stereotypes, which are not significantly correlated across grammatical cases of a pronoun set, nor across models.
Finally, pronoun fidelity is best for traditional pronouns (with no difference, surprisingly, between masculine \textit{er} and feminine \textit{sie}), and models show large quality-of-service differences for more novel, gender-inclusive forms.
Our results highlight the importance of fairness evaluations on morphologically complex languages, and motivate closing these gaps towards inclusive multilingual natural language processing (NLP).
To this end, we release all our code and data.\footnote{\href{https://github.com/TAI-HAMBURG/gruff}{https://github.com/TAI-HAMBURG/gruff}}

\paragraph{Bias statement.}
We take the evidence-based normative position that misgendering is impolite \citep{Tovar+2025+83+94} and harmful \citep{mclemore2018minority}, and that LLM behaviour should respect people's identity, of which pronouns are a part \citep{zimman2019}.

\begin{figure*}[t]
    \centering
    \includegraphics[width=\linewidth]{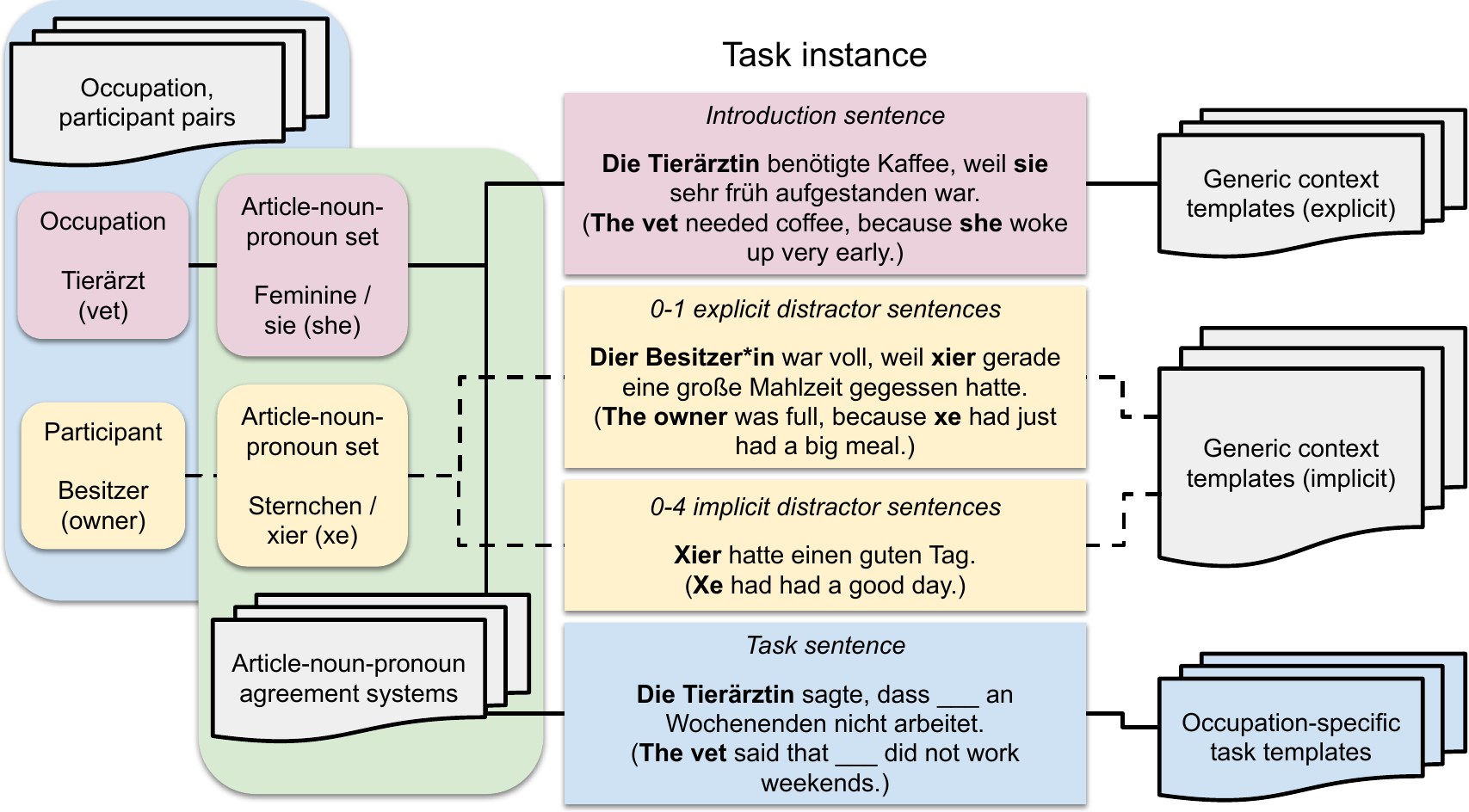}
    \caption{Overview of the creation of each task instance in \texttt{GRUFF}, which involves sampling from occupation-participant pairs (Section \ref{sec:occupations-participants}) and article-noun-pronoun agreement systems (Tables \ref{tab:noun-systems} and \ref{tab:pronoun-sets}) on the left to get the entities being discussed, and sampling from context and task templates (Section \ref{sec:template-creation}) on the right. Assembling these together (Section \ref{sec:template-assembly}) gives task instances like the example in the middle.}
    \label{fig:gruff-data-creation}
\end{figure*}

\section{Related Work}
\paragraph{Gender fairness and identity inclusion in NLP.}
Faithful reproduction of an individual's pronouns falls under research on  gender fairness and identity inclusion, surveyed comprehensively in \citet{stanczak2021survey} and \citet{nemani2024gender}.
Early work in this space showed that NLP systems' predictions and representations reflect and amplify binary gender biases in training data \citep{bolukbasi2016man}, \citet{sun2019mitigating}, and \citet{lu2020gender}.
Beyond English, cross-lingual analyses reveal that gender bias manifests differently across languages and linguistic systems, particularly in languages with grammatical gender \citep{zmigrod-etal-2019-counterfactual,chen2021gender}.
More recent work has expanded the conceptual scope of gender fairness, questioning binary conceptions of gender~\citep{devinney2022theories,cao-daume-iii-2021-toward} and conflating gender with other linguistic categories~\citep{larson2017gender,bartl2025gender}.
Thus, research on identity-inclusive NLP has included on neopronouns and gender-inclusive language \citep{lauscher-etal-2022-welcome,lauscher-etal-2023-em,sobhani2023measuring,waldis-etal-2024-lou}, as part of the emerging area of queer NLP \citep{felkner-etal-2023-winoqueer,weber2026queernlpcriticalsurvey}.

\paragraph{Reasoning with and about pronouns.} A substantial body of research evaluates pronoun handling through coreference benchmarks such as Winograd-style schemas \citep{levesque2012,abdou-etal-2020-sensitivity,emelin-sennrich-2021-wino} and occupational bias datasets \citep{rudinger-etal-2018-gender,zhao-etal-2018-gender,levy-etal-2021-collecting-large}. Earlier work focused primarily on binary pronouns, while more recent studies include singular \textit{they} \citep{baumler-rudinger-2022-recognition} and neopronouns \citep{cao-daume-iii-2021-toward,gautam-etal-2024-winopron,bunzeck-zarriess-2024-slayqa}.
Pronoun agreement is also part of linguistic acceptability judgments, which are often tested with minimal pair datasets such as BLiMP in English \citep{warstadt-etal-2020-blimp-benchmark}, and MultiBLiMP in several other languages \citep{jumelet2025multiblimp10massivelymultilingual}.
\citet{sinha-etal-2023-language} show that such acceptability judgments are context-sensitive, with performance varying under contextual manipulations. %
Context also plays a role in faithful reproduction of individuals' pronouns:
\citet{hossain-etal-2023-misgendered} and \citet{ovalle2023m} evaluate \emph{pronoun fidelity}, but only in simple scenarios with a single individual.
\citet{gautam-et-al-2024-RUFF} consider more complex pronoun fidelity involving multiple individuals, where some are non-adversarial distractors.
\citet{gautam2026} further showed that reasoning models still barely perform above chance in this setting.
However, all these works only consider English.

\paragraph{Multilinguality and pronouns in NLP.}
Multilingual work on pronouns emerged in machine translation, where datasets such as WinoMT help quantify gender bias across language pairs \citep{stanovsky-etal-2019-evaluating,troles-schmid-2021-extending}.
These works often study pronoun preservation when contextual information is required to resolve gender \citep{muller-etal-2018-large,voita-etal-2018-context,fernandes-etal-2023-translation}. \citet{sharma-etal-2022-sensitive} even inject explicit coreference cues to improve pronoun translation.
Other strategies to improve performance include transfer learning and adding relevant information \citep{saunders-byrne-2020-reducing,basta-etal-2020-towards}.
Some works such as \citet{jourdan-2025-fairtranslate} and \citet{bjorklund-devinney-2023-computer} explicitly address neopronouns and gender-inclusive language in French and Swedish, but evaluations of pronoun fidelity remain largely unexplored outside English.

\vspace{0.5em}
\noindent 
\noindent Our work unites these strands by introducing the first large-scale benchmark for pronoun fidelity in German with a range of pronouns, morphological control, and distractor-based tests of robustness. %

\section{Pronoun Fidelity in German}
\label{sec:pronoun-fidelity-task}

The task of pronoun fidelity evaluates the ability of LLMs to consistently reproduce appropriate pronouns in contextualized scenarios, even when intervening distractor sentences introduce competing pronoun references. We modify \citet{gautam-et-al-2024-RUFF}'s original English framework, accounting for German’s rich morphology:
Unlike English, German requires that articles, nouns, and pronouns all agree with each other in grammatical gender (e.g., \emph{der Tierarzt} = \emph{the.MASC.SG vet.MASC.SG}).
Accordingly, for an entity $a$ referred to by their occupation, and an optional second entity $b$ (referred to as a participant), we incorporate gendered articles $a_a$ and $a_b$ that match gender-inflected nouns identifying the entities $e_a$ and $e_b$, as well as the pronouns $p_a$ and $p_b$.
Each task instance consists of:

\vspace{0.5em}
\noindent\textit{Introduction sentence $i(a_a, e_a, p_a)$:} 
Establishes coreference between an entity $a$ (referred to with an article $a_a$ and noun $e_a$, e.g., \emph{der Tierarzt} = \emph{the.MASC.SG vet.MASC.SG}), and a pronoun $p_a$ (e.g., \emph{er} = \emph{he}).

\vspace{0.5em}
\noindent\textit{Distractor sentences $D(a_b, e_b, p_b)$:}
$0$--$5$ sentences that discuss a different entity $b$, either with explicit sentences introducing them with an article $a_b$, noun $e_b$, and pronoun $p_b$, or by implicitly continuing an already-established coreference using only a pronoun $p_b$.

\vspace{0.5em}
\noindent\textit{Task sentence $t(a_a, e_a, p)$:}
Sentence referring to the original entity $a$ with an article $a_a$, noun $e_a$, and a pronoun gap $p$ with an unambiguous reference to the same entity $a$, which must be filled.

\vspace{0.5em}
\noindent Under this general framework, we consider various noun agreement systems and pronoun sets in multiple grammatical cases, as described below.

\section{\texttt{GRUFF} Dataset Creation}
\label{sec:gruff}

\begin{table}[t]
    \centering
    \begin{tabularx}{\columnwidth}{lX}
        \toprule
        \textbf{Noun system} & \textbf{Example} \\
        \midrule
        Masculine & \textit{Der Tierarzt} \\
        Feminine & \textit{Die Tierärztin} \\
        De-e & \textit{De Tierarzte} \\
        Sternchen & \textit{Dier Tierarzt*in} \\
        \bottomrule
    \end{tabularx}
    \caption{The four noun agreement systems we consider for people, including masculine, feminine, and two gender-neutral variants.}
    \label{tab:noun-systems}
\end{table}

\begin{table}[t]
    \centering
    \begin{tabularx}{\columnwidth}{Xcccc}
    \toprule
        \multirow{2}*{\textbf{Case}} & \multicolumn{4}{c}{\textbf{Pronoun}} \\
         & \textbf{Masc.} & \textbf{Fem.} & \multicolumn{2}{c}{\textbf{Neopronouns}} \\
    \midrule
        Nominative & \textit{er} & \textit{sie} & \textit{en} & \textit{xier} \\
        Accusative & \textit{ihn} & \textit{sie} & \textit{en} & \textit{xien} \\
        Genitive & \textit{seine} & \textit{ihre} & \textit{ense} & \textit{xiese} \\
        Dative & \textit{ihm} & \textit{ihr} & \textit{em} & \textit{xiem} \\
    \bottomrule
    \end{tabularx}
    \caption{The four sets of pronouns across four grammatical cases we consider, including masculine and feminine pronouns, and two sets of neopronouns.}
    \label{tab:pronoun-sets}
\end{table}

\begin{table}[t]
    \centering
        \begin{tabularx}{\linewidth}{Xr}
        \toprule
        \bf Data type\phantom{space} & \bf \phantom{space}Number of instances \\
        \midrule
        \rowcolor{grey} \multicolumn{2}{c}{With no context} \\
        Task sentences & $240$ \\
        \midrule
        \rowcolor{grey} \multicolumn{2}{c}{With introductory context} \\
        + $0$ distractors & $3$ x $2,880$ \phantom{$0,00$}(of $9,600$)  \\
        + $1$ distractor\phantom{s} & $3$ x $2,880$ \phantom{$00$}(of $115,200$) \\
        + $2$ distractors & $3$ x $2,880$ \phantom{$00$}(of $460,800$) \\
        + $3$ distractors & $3$ x $2,880$ (of $1,382,400$) \\
        + $4$ distractors & $3$ x $2,880$ (of $2,764,800$) \\
        + $5$ distractors & $3$ x $2,880$ (of $2,764,800$) \\
        \bottomrule
    \end{tabularx}
    \caption{Number of dataset instances in \texttt{GRUFF}. We subsample $3$ sets of $2,880$ instances in each setting with introductory context.}
    \label{tab:data-statistics}
\end{table}

Figure \ref{fig:gruff-data-creation} provides an overview of data creation.

\subsection{Occupations and Participants}
\label{sec:occupations-participants}
We use the same $60$ occupation-participant pairs as in \citet{gautam-et-al-2024-RUFF}, translated to German.
Unlike English, where a single gender-neutral noun (e.g., \emph{vet}) and definite article (\emph{the}) are used regardless of gender, German requires gender-specific articles and nouns.
We consider four forms of articles and nouns (summarized in Table \ref{tab:noun-systems}): Masculine (\textit{der Tierarzt}), feminine (\textit{die Tierärztin}), the de-e system (a German gender-neutral neosystem; \textit{de Tierarzte}), and the gender-inclusive ``Sternchen'' (\textit{dier Tierarzt*in}) \citep{waldis-etal-2024-lou}.

\subsection{Template Creation}
\label{sec:template-creation}

\begin{table*}[t]
    \centering
    \begin{small}
    \begin{tabularx}{\linewidth}{Xlcl}
    \toprule
    \textbf{Model} & \textbf{Size} & \textbf{Quantization} & \textbf{Language} \\
    \midrule
    \rowcolor{grey} \multicolumn{4}{c}{Encoder-only models}\\ 
        \textsc{GBERT-base} \citep{he2023debertav3improvingdebertausing} & $110$M & -- &  German\\
        \textsc{GBERT-large} \citep{he2023debertav3improvingdebertausing} & $340$M & -- & German\\
        \textsc{mBERT-base} \citep{devlin2019bertpretrainingdeepbidirectional} & $178$M & -- & Multilingual\\
        \textsc{XLM-RoBERTa-base} \citep{liu2019robertarobustlyoptimizedbert} & $270$M & -- & Multilingual\\
    \midrule
    \rowcolor{grey} \multicolumn{4}{c}{Decoder-only models}\\
     \textsc{SauerkrautLM-8B} \citep{mayflower2024llama3sauerkrautawq} & $8$B & $4$-bit &German\\
     \textsc{SauerkrautLM-70B} \citep{tresiwalde2024llama3sauerkrautawq} & $70$B & $4$-bit &German\\
     \textsc{Llama-3.3-70B} \citep{llama3modelcard2024} & $70$B & $4$-bit & Multilingual\\
    \bottomrule
    \end{tabularx}
    \end{small}
    \caption{List of models used for our evaluation, along with their architecture, quantization (if applicable), and size.}
    \label{tab:models}
\end{table*}

Every template contains an occupation or participant, as well as a pronoun that matches in gender.
We pair pronoun sets with our four noun systems: Masculine \textit{er}, feminine \textit{sie}, and the neopronouns \textit{en} \citep{vgd-en} and \textit{xier} \citep{heger-xier}.
We consider all four grammatical cases in German, i.e., nominative, accusative, dative, and genitive (see Table \ref{tab:pronoun-sets}).

Introduction and distractor sentences are created from generic, occupation-independent \textbf{context templates} for each pronoun form, while task sentences are created from occupation-specific \textbf{task templates}.
Since translation of English templates can be unnatural or create grammatical constructions that do not measure the phenomena we want, we opt for a mix of translating \texttt{RUFF} templates and creating new templates from scratch.
Creating genitive templates was particularly complex; while English possessive pronouns only inflect for the possessor's number and gender, German possessive pronouns inflect for both the possessor and the possessed object.
We constructed all genitive templates with feminine possessed objects, for a consistent (\textit{-e}) ending across these pronouns and easier evaluation.
In total, we created $240$ occupation-specific task templates ($60$ occupations x $4$ grammatical cases), and $40$ generic context templates ($10$ x $4$ grammatical cases), which are subsequently combined to produce instances.
Of the $240$ task templates, we translated $111$ from English, and created $129$ from scratch.
As for the context templates, we translated $21$, and created $19$ from scratch.

\subsection{Template Assembly}
\label{sec:template-assembly}

Each task template is combined with $10$ introductory templates and filled with $4$ sets of articles, nouns and pronouns, for a total of $9,600$ instances with $0$ distractors.
Distractor sentences are always seeded with a different gender to keep answers unambiguous (see Appendix \ref{sec:template-assembly-math} for details).
With $1$ distractor, there are $115,200$ instances, and with $5$ distractors, there are $2,764,800$.
In total, there are over seven million unique instances (see Table \ref{tab:data-statistics}).
As it is computationally infeasible to evaluate with all of these, for each split of $0$--$5$ distractors, we randomly downsample $3$ sets of $2,880$, with equal representation of each of the $4$ pronoun sets.

\subsection{Human Validation}
Three annotators (including one author) independently validated all context templates for grammatical correctness.
They also filled in the pronoun with $100$\% accuracy for a subset of $600$ final instances from \texttt{GRUFF} ($100$ randomly sampled instances for each of $0$--$5$ distractor sentences).
Further details on annotation and annotators are in Appendix \ref{sec:annotation-details}.

\section{Experimental Setup}

\begin{figure*}[t]
    \centering
    \begin{subfigure}[t]{\textwidth}
        \includegraphics[width=0.32\linewidth]{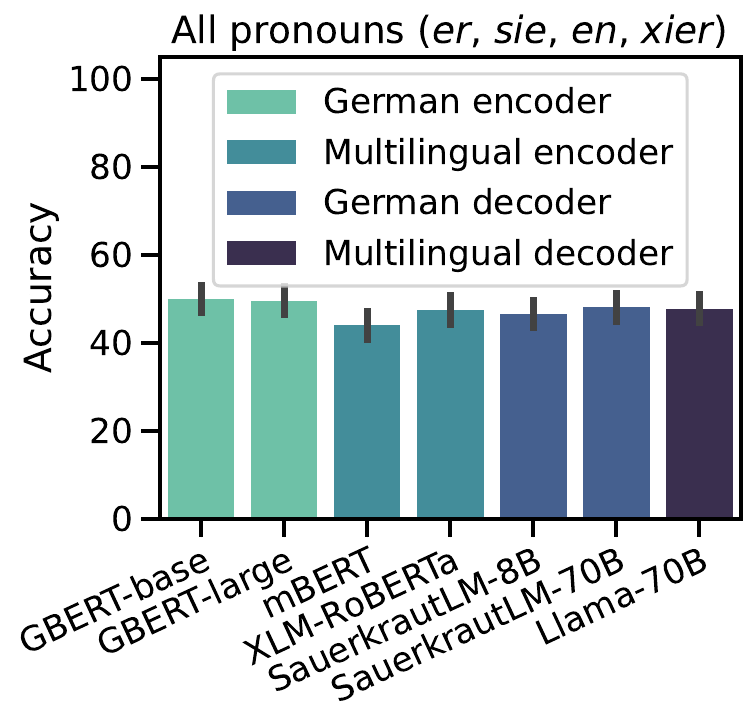}
        \includegraphics[width=0.32\linewidth]{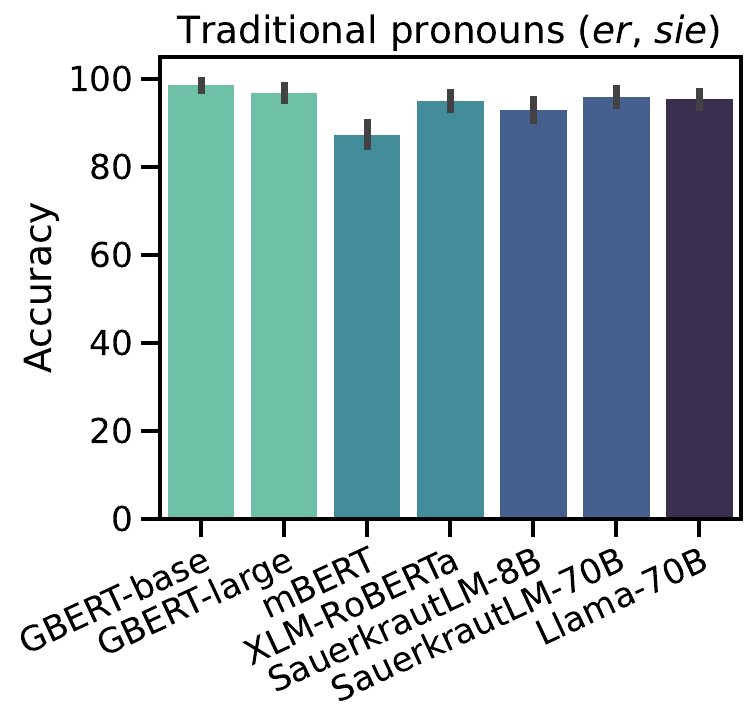}
        \includegraphics[width=0.32\linewidth]{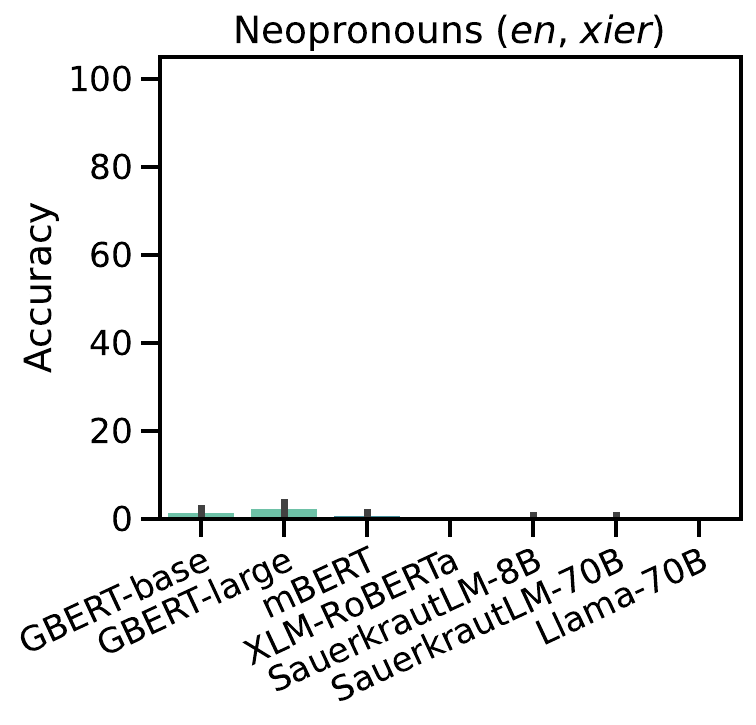}
        \caption{With \textit{no} context (just grammatical gender agreement)}
        \label{fig:no-context-agreement}
    \end{subfigure}
    \begin{subfigure}[t]{\textwidth}
        \includegraphics[width=0.32\linewidth]{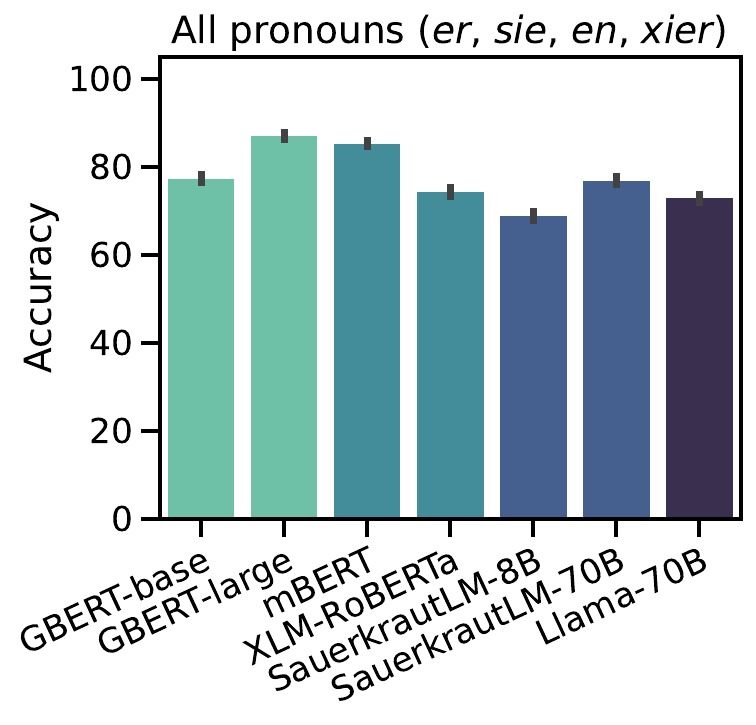}
        \includegraphics[width=0.32\linewidth]{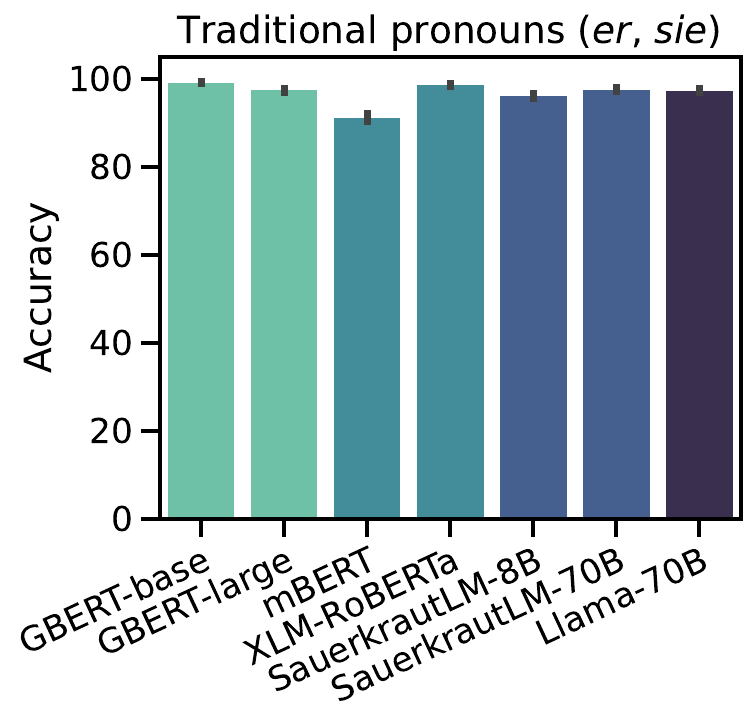}
        \includegraphics[width=0.32\linewidth]{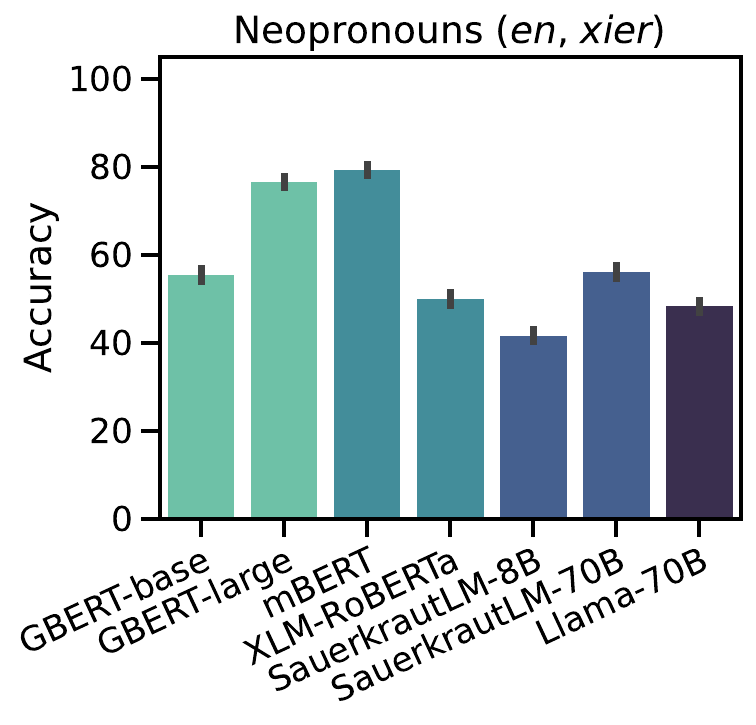}
        \caption{With context explicitly specifying a pronoun}
        \label{fig:with-context-agreement}
    \end{subfigure}
    \caption{Model accuracy on grammatical gender agreement with and without context. Without context, models are near-perfect for traditional masculine and feminine pronouns, but cannot produce any neopronoun. With an explicitly specified pronoun, models can reproduce neopronouns and maintain performance on traditional pronouns.}
    \label{fig:agreement}
\end{figure*}

\subsection{Models}
We select four encoder-only and three decoder-only models that support the German language, covering a range of sizes and pre-training languages, summarized in Table \ref{tab:models}.
We aim to study the effects of model architecture and the differences between German-only and multilingual models.

\subsection{Evaluation}
To enable consistent, automatic evaluation across model architectures, we use a forced-choice setting where models select one of four pronoun options.
Concretely, we fill the pronoun gap in $t(a_a, e_a, p)$ with each pronoun of the appropriate case.
We then compute average sequence probabilities, using log-likelihoods for decoder-only models, and pseudo-log-likelihoods for encoder-only models \citep{salazar-etal-2020-masked}.
We average rather than sum probabilities, as the number of tokens differs substantially across variants.
Finally, the highest-likelihood option is taken as the model's answer.
We opt for probability- rather than generation-based evaluation as we are interested in LLM capabilities as models of language \citep{subramonian2025agree}.

Human accuracy of $100$\% represents the performance ceiling that language models could theoretically achieve.
\citet{gautam-et-al-2024-RUFF} use a $25$\% baseline, the probability of choosing between four pronouns at random.
We do not indicate this baseline as it is unrealistic; models overpredict and better resolve traditional pronouns over neopronouns.

\section{Grammatical Agreement}
\label{sec:agreement}

\begin{figure*}[t]
    \includegraphics[width=\linewidth]{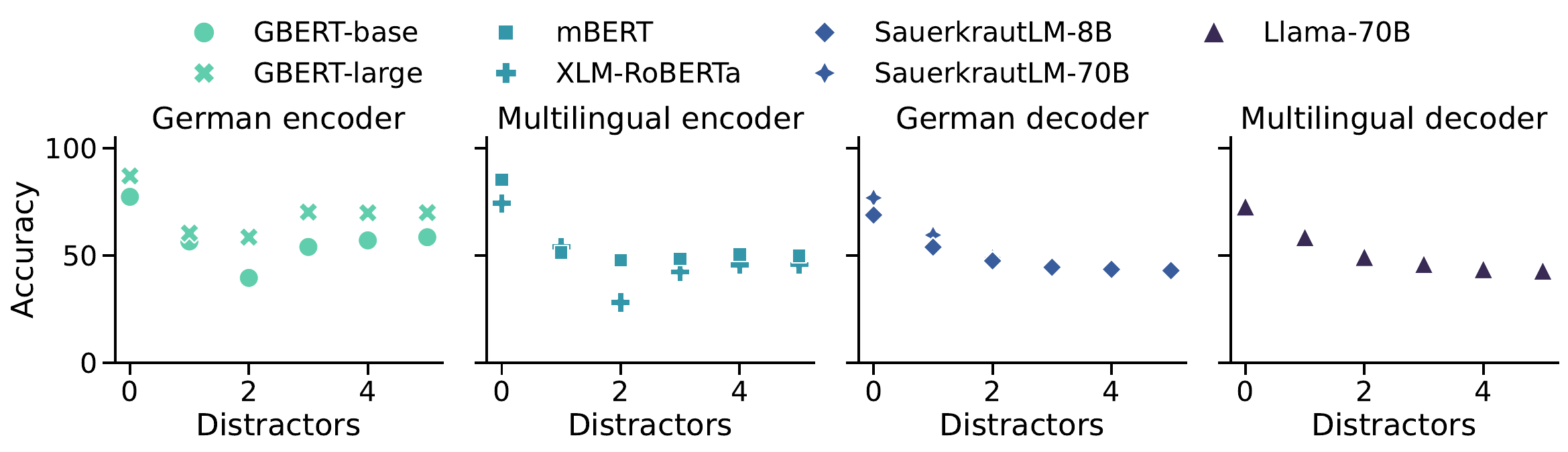}
    \caption{Accuracy at pronoun fidelity with an introductory sentence and $0$--$5$ distractors across different models.}
    \label{fig:distractor-accuracy}
\end{figure*}

\begin{table*}
    \small
    \begin{tabularx}{\linewidth}{Xrrrrr}
    \toprule
    \multirow{2}{*}{\textbf{Model}} & \multicolumn{5}{c}{\textbf{Percentage of errors that are due to distraction}} \\
    & \multicolumn{1}{c}{\textbf{1 distractor}} & \multicolumn{1}{c}{\textbf{2 distractors}} & \multicolumn{1}{c}{\textbf{3 distractors}} & \multicolumn{1}{c}{\textbf{4 distractors}} & \multicolumn{1}{c}{\textbf{5 distractors}}  \\
    \midrule
    \rowcolor{grey} \multicolumn{6}{c}{Encoder-only models} \\ 
    \textsc{GBERT-base} & $85.59 \pm 0.62$ & \dag$94.72 \pm 0.21$ & $86.60 \pm 0.57$ & $83.87 \pm 0.71$ & $82.82 \pm 0.40$ \\
    \textsc{GBERT-large} & $85.56 \pm 0.14$ & \dag$86.65 \pm 0.37$ & $78.87 \pm 0.91$ & $79.41 \pm 0.54$ & $79.00 \pm 0.93$ \\
    \textsc{mBERT} & \dag$86.19 \pm 1.26$ & $86.07 \pm 0.24$ & $83.65 \pm 0.26$ & $82.70 \pm 0.22$ & $81.71 \pm 0.58$ \\
    \textsc{XLM-RoBERTa} & $80.98 \pm 0.79$ & \dag$93.36 \pm 0.09$ & $87.74 \pm 0.71$ & $85.25 \pm 0.29$ & $84.44 \pm 0.11$ \\
    \midrule
    \rowcolor{grey} \multicolumn{6}{c}{Decoder-only models}\\ 
    \textsc{SauerkrautLM-8B} & $56.92 \pm 0.14$ & $65.99 \pm 0.46$ & $69.52 \pm 0.24$ & $69.61 \pm 0.57$ & \dag$69.77 \pm 0.49$ \\
    \textsc{SauerkrautLM-70B} & $75.21 \pm 0.43$ & $82.70 \pm 0.55$ & $83.93 \pm 0.39$ & $84.70 \pm 0.34$ & \dag$85.47 \pm 0.40$ \\
    \textsc{Llama-70B} & $72.39 \pm 0.86$ & $79.95 \pm 1.06$ & $83.32 \pm 0.55$ & $83.26 \pm 1.09$ & \dag$83.56 \pm 0.94$ \\
    \bottomrule
    \end{tabularx}
    \caption{All model errors are mostly due to distraction. With more distractors, encoder-only models are less distracted, while decoder-only models are more distracted. \dag: Highest distraction error in each row.}
    \label{tab:distraction-errors}
\end{table*}

As the German language features rich morphological markings for grammatical gender, we can expect that a language model that is good at modelling German will show grammatical agreement in the absence of any introductory sentence, i.e., masculine nouns should take masculine pronouns, feminine nouns should take feminine pronouns, and de-e and Sternchen nouns can correctly take either \textit{en} or \textit{xier}.\footnote{As there is no consensus on which neopronoun sets map to which gender-neutral noun-article sets, we evaluate generously, expecting neopronouns to be paired with \textit{any} gender-neutral article and noun over explicitly masculine or feminine variants. This is in line with what is acceptable to native speakers and what a strong LLM should be able to reproduce.}
We can already measure this in the no-context case, unlike in English, and then investigate whether explicitly specifying a pronoun \textit{improves} agreement.

As Figure \ref{fig:agreement} shows, \textbf{models show near-perfect grammatical agreement for traditional pronouns}, independent of context.
In contrast, \textbf{models show near-zero agreement for neopronouns without explicit context} which includes \textit{xier} or \textit{en} pronouns.
The stark difference between Figures \ref{fig:no-context-agreement} and \ref{fig:with-context-agreement} for neopronouns indicates that explicitly conditioning on them improves pronoun fidelity.

The patterns and raw performance scores are consistent across models with different pre-training language distributions, as well as across architectures.
This corroborates \citeauthor{gautam-et-al-2024-RUFF}'s \citeyearpar{gautam-et-al-2024-RUFF} findings that encoder-only models are far better than decoder-only models at comparable scales, as the decoder-only models we consider are also $80$--$700$ times larger than the encoder-only ones.

\section{Robustness to Distractors}
\label{sec:distractors}

Next, we investigate how non-adversarial distractors (in the form of a second entity with different pronouns) affect model performance.
In English, performance drops dramatically with just one distractor.
However, German's encoding of gender agreement is likely to make models more robust.

In Figure \ref{fig:distractor-accuracy}, we see that \textbf{distractors affect performance dramatically, despite the additional signal of grammatical gender}.
These trends are identical for traditional pronouns and neopronouns, except for an overall downward shift in performance with neopronouns.
Patterns by architecture are nearly identical to \citet{gautam-et-al-2024-RUFF}, likely due to bidirectional attention and evaluation with pseudo log likelihoods for encoder-only models:
As in English, additional distractors worsen decoder-only model performance, but encoder-only models recover marginally.
German-only and multilingual models largely follow similar trends.

The vast majority of errors happen when models repeat the distractor pronoun, as Table \ref{tab:distraction-errors} shows.
Interestingly, \textbf{encoder-only models are most distracted with \textit{two} distractors}, rather than one, as in English.
We posit that this is a direct result of more gender marking in German, which is a stronger signal to models, enabling robust pronoun fidelity when only one distractor is present.

\section{Bias and Stereotypes}
\label{sec:bias}

\begin{figure*}[t]
    \centering
    \includegraphics[scale=0.4]{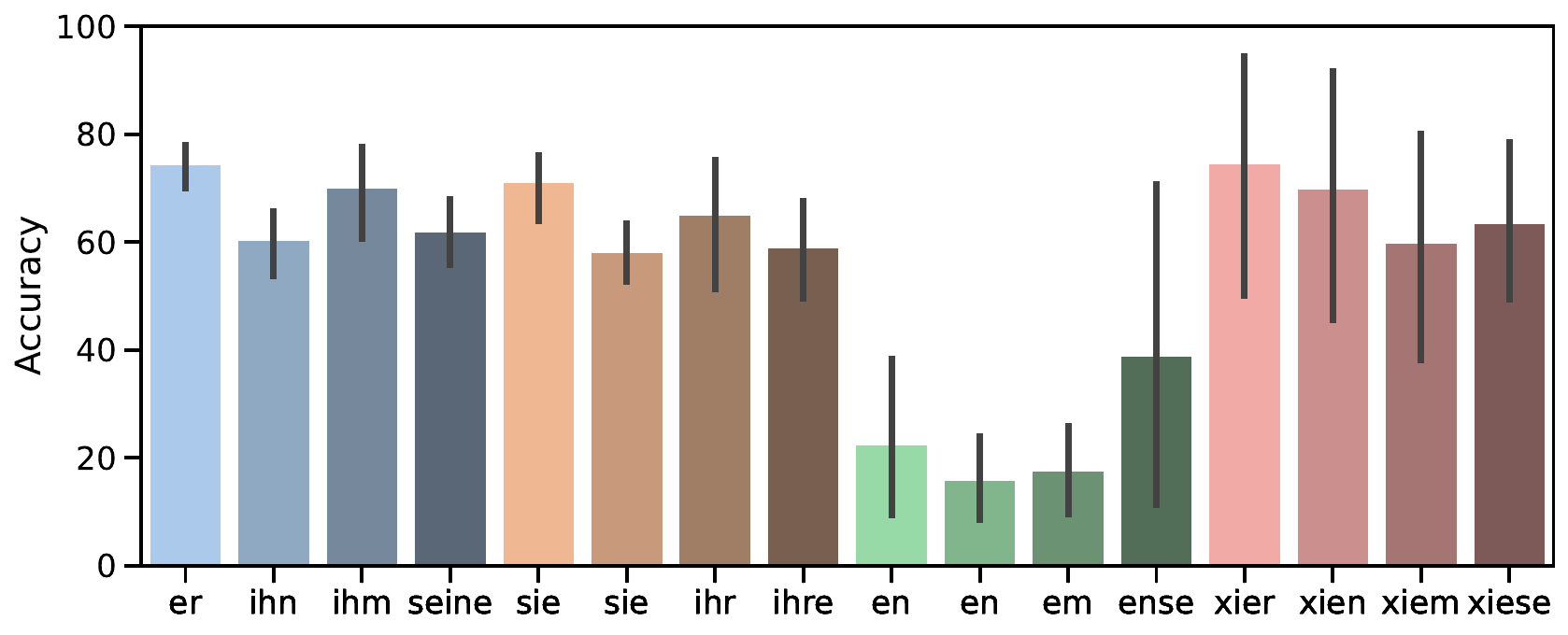}
    \caption{Pronoun fidelity split by pronoun set and by grammatical case. Overall models are best at reusing \textit{er}, \textit{sie}, and the neopronoun \textit{xier}. Performance with a different neopronoun, \textit{en}, is very low.}
    \label{fig:pronoun-fidelity-by-pronoun}
\end{figure*}

\begin{figure*}[t]
    \centering
    \includegraphics[width=\linewidth]{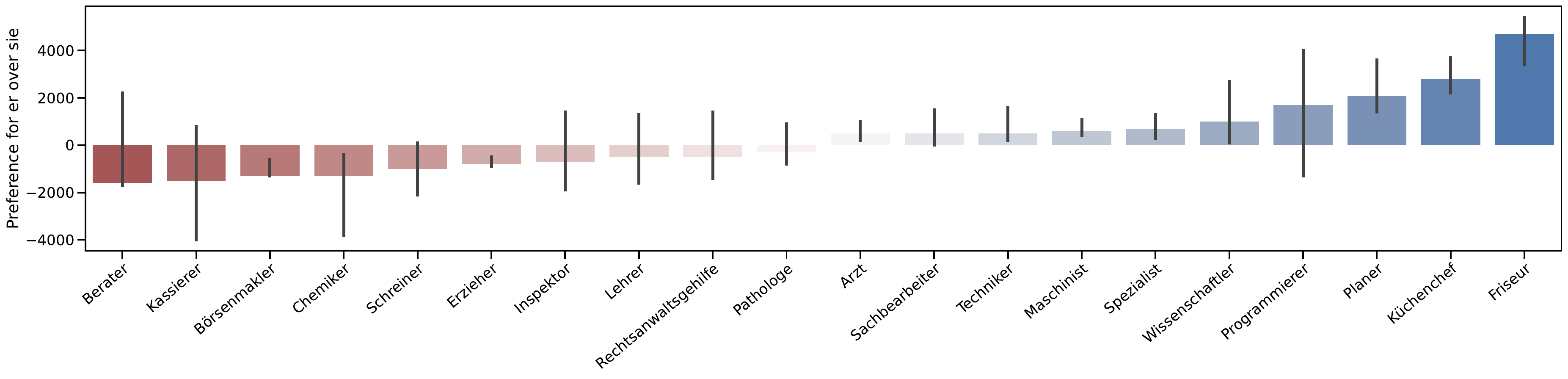}
    \caption{Top $10$ occupations over-resolved to accusative pronouns \textit{ihn} (masculine) on the right, versus \textit{sie} (feminine) on the left, in a pattern indicative of stereotyping. Results with other grammatical cases are shown in Appendix \ref{sec:additional-results}.}
    \label{fig:stereotyping-case-accusative}
\end{figure*}

\begin{figure}[t]
    \centering
    \includegraphics[width=\linewidth]{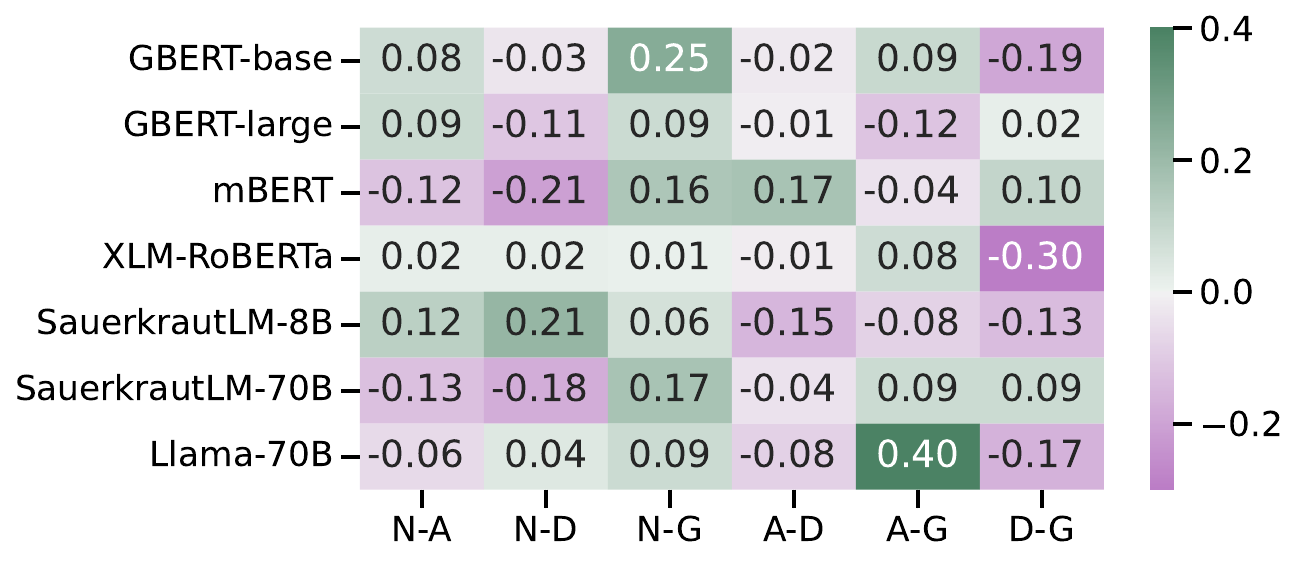}
    \caption{Spearman's correlation between the stereotypical biases across different cases, on a per model basis. N: Nominative, A: Accusative, D: Dative, G: Genitive, and * indicates statistical significance ($\alpha=0.05$). Correlations are generally low and even the higher values are not significant.}
    \label{fig:stereotyping-case}
\end{figure}

\begin{figure}[t]
    \centering
    \includegraphics[width=\linewidth]{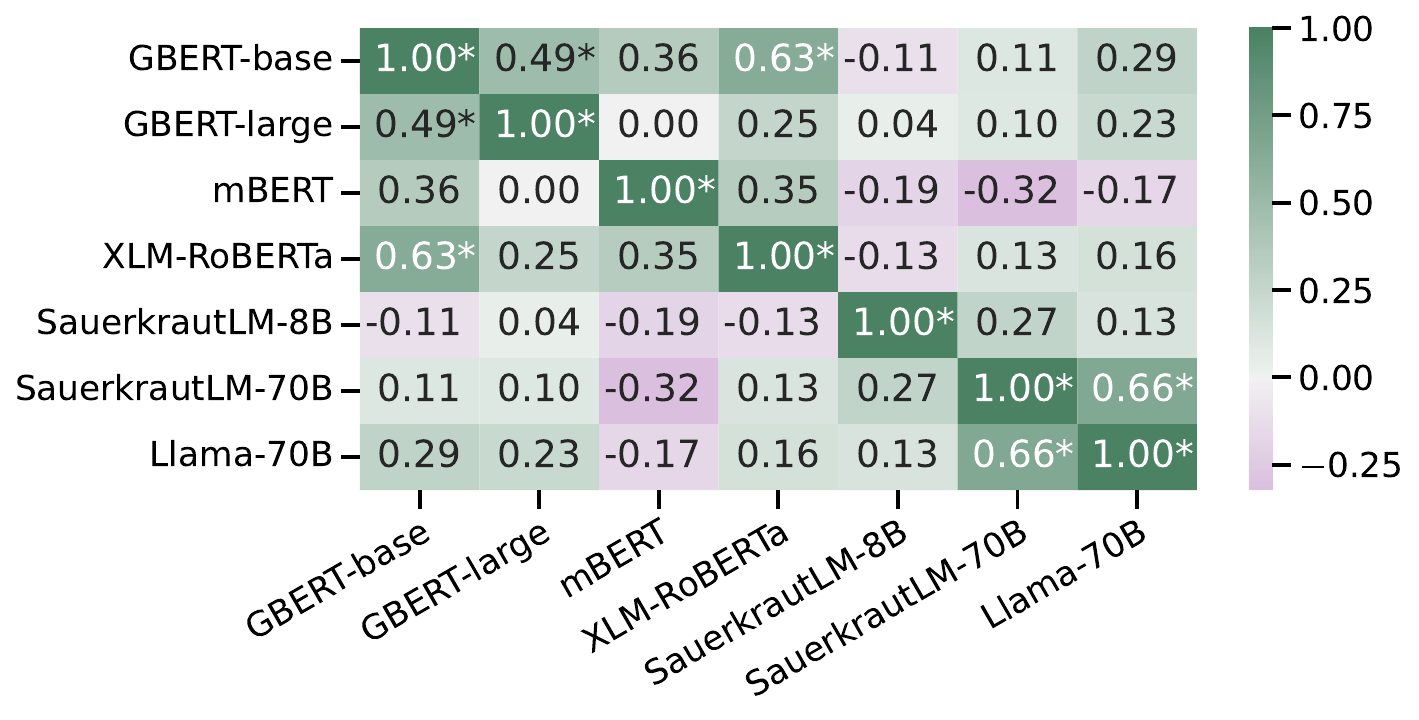}
    \caption{Spearman's correlation between the stereotypical biases of models for accusative pronouns. * indicates statistical significance ($\alpha=0.05$).}
    \label{fig:stereotyping-models-accusative}
\end{figure}

The previous sections consider pronoun fidelity from the perspectives of grammatical agreement and robustness.
In this section, we turn to concerns of biases and stereotyping.
In English, \citet{gautam-et-al-2024-RUFF} report statistically significant differences in pronoun fidelity with \textit{he/him/his} pronouns compared to \textit{she/her/her}, and a further drop in performance for singular \textit{they} and the neopronoun \textit{xe/xem/xyr}.
We now investigate whether similar quality-of-service differentials \citep{cao-daume-iii-2021-toward} appear with German.

Across all settings with an explicit introduction (i.e., with $0$--$5$ distractors), pronoun fidelity is $66.52 \pm 10.59$ for \textit{er/ihm} pronouns, $63.18 \pm 12.58$ for \textit{sie/ihr}, $23.49 \pm 26.05$ for \textit{en/em}, and $66.74 \pm 29.54$ for \textit{xier} pronouns.
With a Wilcoxon signed-rank test ($\alpha=0.05$), there are \textbf{no statistically significant differences in performance averages between \textit{er}, \textit{sie}, and \textit{xier} pronouns} (the last of which has a much larger variance), while \textbf{\textit{en} is significantly worse}.\footnote{In these results, \textit{en} pronouns are mapped exclusively to de-e articles and nouns (and \textit{xier} to the more common Sternchen form), but this pattern holds even when the mappings are swapped, as we show in Appendix \ref{sec:en-xier-nouns}.}
The parity in \textit{er} and \textit{sie} performance is likely because gender information from the noun provides an additional and repeated cue for improved pronoun fidelity in German.
This could also explain why there are no statistically significant differences in performance between different grammatical cases of the same pronoun set (see Figure \ref{fig:pronoun-fidelity-by-pronoun}), unlike in English \citep{gautam-etal-2024-winopron}.
Taken together, these results show that pronoun fidelity in German, while not robust to distractors, is characterized by less bias than English precisely because of more morphological gender marking.
\citet{rodriguez-etal-2025-colombian} finds similar results with Spanish, despite hypothesizing that more gender marking would result in more bias.

Finally, we study occupational biases, comparing how much models over-resolve a particular pronoun to to an occupation, even overriding grammatical gender signals from articles and nouns.
We exclude neopronouns, as their behaviour is less interpretable in terms of social gender stereotypes and less likely to be learned from data due to sparsity.
We consider all settings with an introduction and any number of distractors.
As Figure \ref{fig:stereotyping-case-accusative} shows, \textit{Friseur} (hairdresser), \textit{Planer} (planner), and \textit{Küchenchef} (chef) are overresolved to masculine \textit{ihn}, while \textit{Berater} (counsellor), \textit{Kassierer} (cashier), and \textit{Börsenmakler} (broker), are overresolved to \textit{sie}.

As this analysis produces a ranking, we can use rank correlation metrics to understand whether occupational biases of models with different grammatical cases are similar, and whether the occupational biases of different \textit{models} are similar.
We use Spearman's $\rho$ and report test statistics as well as significance ($\alpha=0.05$, with Bonferroni correction).
Similar to \citet{gautam-etal-2024-winopron}, individual models show \textbf{no significant correlation between stereotypical biases across different grammatical cases of the same pronoun set}, as shown in Figure \ref{fig:stereotyping-case}.
Models, however, tend to be slightly correlated with each other (see Figure \ref{fig:stereotyping-models-accusative}), particularly when they have a similar architecture, e.g., \textsc{GBERT-base} and \textsc{-large}, and \textsc{SauertkrautLM-70B} and \textsc{Llama-70B}.
However, this does not always hold, as \textsc{SauerkrautLM-8B} and \textsc{-70B} are very weakly correlated.
Overall, we see that German-language occupational biases are substantially different from case to case, and model to model, but that the overall issues of gender bias are substantially different in a gender-marked language such as German, compared to English.

In sum, quality-of-service differentials in German pronoun fidelity are low with no masculine-feminine difference (unlike English). Stereotyping is both weak and weakly correlated across settings.

\section{Discussion and Conclusion}

The \texttt{GRUFF} dataset that we propose enables us to evaluate a range of language models with the lenses of grammatical agreement, robustness to discourse complexity (in the form of distractor sentences about other entities), as well as biases and stereotypes.
Although our work is inspired by the English \texttt{RUFF} dataset, these issues are fundamentally different in a morphologically rich language like German which marks gender on articles, nouns, and pronouns.
We show that these linguistic differences result in higher and more robust pronoun fidelity.
Somewhat counterintuitively, more explicit marking of gender results in less stereotyping, and more balance in pronoun fidelity across grammatical genders.
Our results thus highlight the importance of fairness research that goes beyond English.
    
Nevertheless, performance with neopronouns and gender-inclusive articles and nouns remains mixed, particularly with \textit{en} pronouns, the recommendation of the society for gender-neutral German, \citeauthor{vgd-en}.
Closing these gaps should be an important priority for future work on inclusive multilingual NLP.

\section*{Limitations}

The primary limitation of our study is its ecological validity, as we use researcher-constructed synthetic templates rather than natural discourse, and a forced-choice evaluation setting with language model probabilities.
For the questions we are asking about how LLMs model agreement and how pronouns of different discourse entities interact, this is the only setting that provides sufficient control, as in \cite{gautam-et-al-2024-RUFF}, but results with generations or real-world data may differ \citep{subramonian2025agree}.
Our evaluation is also limited to encoder-only and decoder-only models, and we do not consider encoder-decoder architectures, models with extensive post-training (e.g., chat models). or reasoning models.

As part of formalizing \texttt{GRUFF}, we also made certain decisions about which gender-neutral article, noun, and pronoun forms to include, based on recommendations from advocacy groups and members of the German non-binary community that have become widely adopted.
However, there is no social consensus on gender-fair or gender-neutral forms in German at the moment.
Some individuals opt for no pronouns in German \citep{sawall-2024-thesis}, and some use \textit{dey} pronouns \citep{huck-2021-thesis}, which are some examples of strategies we do not consider.
Furthermore, \textit{English} plays a large role in the linguistic practices of the German queer community \citep{zieglmeier-2023-queer-english,sawall-2024-thesis}.
Finally, we do not consider how queer language interacts with other aspects of identity (e.g., names, ethnic identity), and discourse (e.g., domain).

\section*{Ethics Statement}

As our work touches on socially sensitive language phenomena, there are a number of ethical implications.
Our results show how LLMs can harm people, particularly non-binary individuals and users of gender-neutral forms more broadly.
If systems that display poor pronoun fidelity are deployed in high-impact domains like education and public services, such errors may reinforce exclusion, reduce trust, and contribute to discriminatory outcomes.
The only potential risk we see of our work is that our benchmark assigns certain pronouns as ``correct,'' which do not represent the large variety of pronoun practices people employ in German and beyond.
\texttt{GRUFF} should therefore be treated a diagnostic benchmark rather than a normative authority on language use.
Scores should be interpreted as indicators of model behaviour rather than definitive judgments about people's identities or acceptable language in all contexts.
Responsible use of \texttt{GRUFF} requires transparent reporting, context-aware interpretation of results, and only using the benchmark for evaluation rather than training, due to the possibility of models exploiting shallow heuristics, explained in \cite{gautam-et-al-2024-RUFF}.

\bibliography{custom}

\appendix

\begin{table*}[t]
    \centering
    \small
    \begin{tabular*}{\textwidth}{@{\extracolsep{\fill}}crcrcrcrcrcrc@{}r}
    \toprule
    \textbf{\shortstack{Number of\\distractors}} &
    $\mathbf{t(a_a, e_a, p)}$ &
    &
    $\mathbf{i(a_a, e_a, p_a)}$ &
    &
    $\mathbf{C}$ &
    &
    $\mathbf{d(a_b, e_b, p_b)}$ &
    &
    $\mathbf{p_b}$ &
    &
    $\mathbf{P}$ &
    &
    \textbf{\shortstack{Total\\instances}} \\
    \midrule
    $0$ & $240$ & $\times$ & $10$ & $\times$ & $4$ &  & &  &  &  &  & $=$ & $9,600$ \\
    $1$ & $240$ & $\times$ & $10$ & $\times$ & $4$ & $\times$ & $4$ & $\times$ & $3$ &  &  & $=$ & $115,200$ \\
    $2$ & $240$ & $\times$ & $10$ & $\times$ & $4$ & $\times$ & $4$ & $\times$ & $3$ & $\times$ & $4$ & $=$ & $460,800$ \\
    $3$ & $240$ & $\times$ & $10$ & $\times$ & $4$ & $\times$ & $4$ & $\times$ & $3$ & $\times$ & $12$ & $=$ & $1,382,400$ \\
    $4$ & $240$ & $\times$ & $10$ & $\times$ & $4$ & $\times$ & $4$ & $\times$ & $3$ & $\times$ & $24$ & $=$ & $2,764,800$ \\
    $5$ & $240$ & $\times$ & $10$ & $\times$ & $4$ & $\times$ & $4$ & $\times$ & $3$ & $\times$ & $24$ & $=$ & $2,764,800$ \\
    \bottomrule
    \end{tabular*}
    \caption{Computation of the total number of \texttt{GRUFF} instances per distractor setting. Total instances are obtained by multiplying the factors in each row from left to right. Here, $t(a_a, e_a, p)$ denotes task templates, $i(a_a, e_a, p_a)$ explicit introduction context templates, $C$ grammatical cases, $d(a_b, e_b, p_b)$ explicit distractor context templates, $p_b$ distractor pronoun options, and $P$ implicit distractor sentence permutations.}
    \label{tab:dataset_instances}
\end{table*}

\section{Further Details on Template Assembly}
\label{sec:template-assembly-math}

The number of dataset instances in Table \ref{tab:data-statistics} is calculated as follows, where the symbols are the components of the German pronoun fidelity task described in Section \ref{sec:pronoun-fidelity-task}:
\begin{multline*}
    \text{$0$ distractors: $40$ $i(\textcolor{forestgreen}{a_a},\textcolor{forestgreen}{e_a}, \textcolor{forestgreen}{p_a})$ * $240$ $t(\textcolor{forestgreen}{a_a},\textcolor{forestgreen}{e_a}, \textcolor{black}{p})$}
\end{multline*}
\begin{multline*}
    \text{$1$ distractor: $40$ $i(\textcolor{forestgreen}{a_a},\textcolor{forestgreen}{e_a}, \textcolor{forestgreen}{p_a})$ * $240$ $t(\textcolor{forestgreen}{a_a},\textcolor{forestgreen}{e_a}, \textcolor{black}{p})$} \\ \text{* $12$ $d(\textcolor{red}{{a_b}},\textcolor{red}{e_b}, \textcolor{red}{p_b})$}
\end{multline*}
\begin{multline*}
\text{$2$--$5$ distractors: $40$ $i(\textcolor{forestgreen}{a_a},\textcolor{forestgreen}{e_a}, \textcolor{forestgreen}{p_a})$ * $240$ $t(\textcolor{forestgreen}{a_a},\textcolor{forestgreen}{e_a}, \textcolor{black}{p})$} \\ \text{* $12$ $d(\textcolor{red}{{a_b}},\textcolor{red}{e_b}, \textcolor{red}{p_b})$ * $P$ * $d(\textcolor{red}{p_b})$}
\end{multline*}

For the templates with $2$--$5$ distractors, the number of permutations $(P)$ is calculated as follows:
\begin{multline*}
\text{$2$ distractors:} \quad P(4,1) = \frac{4!}{(4-1)!} = \frac{24}{6} = 4
\end{multline*}
\begin{multline*}
\text{$3$ distractors:} \quad P(4,2) = \frac{4!}{(4-2)!} = \frac{24}{2} = 12
\end{multline*}
\begin{multline*}
\text{$4$ distractors:} \quad P(4,3) = \frac{4!}{(4-3)!} = \frac{24}{1} = 24
\end{multline*}
\begin{multline*}
\text{$5$ distractors:} \quad P(4,4) = \frac{4!}{(4-4)!} = \frac{24}{1} = 24
\end{multline*}

The number of dataset instances for $4$ and $5$ distractors is identical because of the permutation calculation shown above.
Table~\ref{tab:dataset_instances} shows the calculations for the dataset construction for each distractor setting in more detail.

\section{Human Validation}
\label{sec:annotation-details}

The validation was designed to ensure inter-annotator reliability through identical task assignment.
Data was provided to annotators in Google Sheets for automatic validation.
$600$ filled context templates were provided in one sheet.
A second sheet contained $600$ task dataset instances ($100$ randomly sampled instances each for settings with $0$--$5$ distractor sentences).
Annotators achieved $100$\% accuracy in reproducing the correct pronoun on these instances.
Below we present annotator demographics as well as the original annotation instructions (in German).

\subsection{Annotator Demographics}

Three annotators who are German native speakers participated in the human validation, one of whom is an author of this work.
The two other annotators were students recruited from the university community; annotators 1 and 2 were not paid, and annotator 3 completed the annotations as part of his work contract as a student assistant.

\paragraph{Annotator 1 (Author)}
\begin{itemize}
  \item \textbf{Age:} $25$
  \item \textbf{Gender:} Male
  \item \textbf{Special knowledge/experience with German gender-fair language and neopronouns:} High. Annotator 1 is the lead author of the work and conducted extensive research on this topic.
\end{itemize}

\paragraph{Annotator 2}
\begin{itemize}
  \item \textbf{Age:} $26$
  \item \textbf{Gender:} Female
  \item \textbf{Special knowledge/experience with German gender-fair language and neopronouns:} No prior experience before validating the dataset.
\end{itemize}

\paragraph{Annotator 3}
\begin{itemize}
  \item \textbf{Age:} $27$
  \item \textbf{Gender:} Male
  \item \textbf{Special knowledge/experience with German gender-fair language and neopronouns:} Prior professional work involving German gender-fair language and neopronoun usage.
  \end{itemize}

\subsection{Guidelines}

Here we show the German annotation guidelines as originally provided to the annotators.
They are heavily based on the English annotation guidelines in \citet{gautam-et-al-2024-RUFF}.

\subsection*{Anmerkungsinstruktionen}\label{sec:anmerkungs-instruktionen}
\subsubsection*{Instruktionen zu Aufgabe 1}
Zusammen mit diesem Annotationsprotokoll haben Sie den Link zu einem Google Sheet erhalten. Das Sheet enth\"alt 2 Datenspalten und 2 Aufgabenspalten mit zuf\"alligen Daten. Die Datenspalten bestehen aus:
\begin{itemize}
    \item S\"atzen, die Sie auf grammatikalische Korrektheit bewerten sollen;
    \item Fragen zu Pronomen in den S\"atzen, die Sie beantworten sollen
\end{itemize}
Bitte seien Sie pr\"azise in Ihren Zuordnungen und \"andern Sie nicht die Reihenfolge der Daten. Die Spalten haben eine integrierte Datenvalidierung und wir werden weitere Tests durchf\"uhren, um eine konsistente Annotation zu \"uberpr\"ufen.

\subsubsection*{Grammatikalische Korrektheit}
In der Spalte "Grammatikalisch korrekt?" geben Sie bitte Ihre Bewertung des Satzes nach Standard-Deutsch ein. Die Anmerkungsoptionen sind:
\begin{itemize}
    \item \textbf{Grammatikalisch korrekt} (f\"ur fl\"ussige, syntaktisch g\"ultige und semantisch plausible S\"atze)
    \item \textbf{Grammatikalisch nicht korrekt} (f\"ur S\"atze mit Tippfehlern, grammatikalischen Problemen, oder wenn der Satz eine Situation beschreibt, die keinen Sinn ergibt oder einfach seltsam klingt)
    \item \textbf{Nicht sicher} (wenn Sie nicht sicher sind, ob der Satz eindeutig grammatikalisch korrekt oder grammatikalisch nicht korrekt ist)
\end{itemize}
Beispiele:
\begin{itemize}
    \item \textit{Der Psychologe f\"uhlte sich erholt; seine Nacht verlief ruhig.}\\
    \=\> grammatikalisch korrekt
    \item \textit{Der Fahrer sagte dem Fahrgast er kann die Fahrt bar bezahlen aber vergaß das Wechselgeld.}\\
    \=\> grammatikalisch nicht korrekt (wegen fehlender Kommasetzung, nicht korrekter deutscher Grammatik oder unklarer Bez\"uge)
\end{itemize}

\begin{table*}[t]
    \centering
    
    \begin{tabularx}{\linewidth}{lX}
    \toprule
    \textbf{Model} & \textbf{HuggingFace Identifier} \\
    \midrule
    \rowcolor{grey} \multicolumn{2}{c}{Encoder-only models}\\ 
        \textsc{GBERT-base} & \href{https://huggingface.co/deepset/gbert-base}{deepset/gbert-base} \\
        \textsc{GBERT-large} & \href{https://huggingface.co/deepset/gbert-large}{deepset/gbert-large} \\
        \textsc{mBERT} & \href{https://huggingface.co/google-bert/bert-base-multilingual-cased}{google-bert/bert-base-multilingual-cased} \\
        \textsc{XLM-RoBERTa} & \href{https://huggingface.co/FacebookAI/xlm-roberta-base}{FacebookAI/xlm-roberta-base} \\
    \midrule
    \rowcolor{grey} \multicolumn{2}{c}{Decoder-only models}\\
     \textsc{SauerkrautLM-8B} & \href{https://huggingface.co/mayflowergmbh/Llama-3-SauerkrautLM-8b-Instruct-GGUF}{mayflowergmbh/Llama-3-SauerkrautLM-8b-Instruct-GGUF} \\
     \textsc{SauerkrautLM-70B} & \href{https://huggingface.co/tresiwalde/Llama-3-SauerkrautLM-70b-Instruct-AWQ}{tresiwalde/Llama-3-SauerkrautLM-70b-Instruct-AWQ} \\
     \textsc{Llama-3.3-70B} & \href{https://huggingface.co/casperhansen/llama-3.3-70b-instruct-awq}{casperhansen/llama-3.3-70b-instruct-awq} \\
    \bottomrule
    \end{tabularx}
    
    \caption{List of models along with their HuggingFace identifiers.}
    \label{tab:models-huggingface}
    
\end{table*}

\subsubsection*{Fragen zu Pronomen}
Jeder Satz enthält ein Pronomen, und die Spalte ``Frage'' fragt, ob es sich auf eine im Satz erw\"ahnte Person mit der jeweiligen Berufsbezeichnung bezieht oder nicht. Die Anmerkungsoptionen sind:
\begin{itemize}
    \item \textbf{ja} (wenn sich das Pronomen auf die Person mit der genannten Berufsbezeichnung bezieht)
    \item \textbf{nein} (wenn sich das Pronomen nicht auf die Person mit der genannten Berufsbezeichnung bezieht)
    \item \textbf{nicht sicher} (wenn Sie nicht sicher sind, ob sich das Pronomen auf die Person mit der genannten Berufsbezeichnung bezieht)
\end{itemize}
Beispiele: 
\begin{itemize}
    \item \textit{Der Psychologe f\"uhlte sich erholt; ihre Nacht verlief ruhig.}\\
    Frage: Bezieht sich ihre auf die Person mit der Berufsbezeichnung Psychologe?\\
    \=\> Nein
    \item \textit{Die Ausbilderin ben\"otigte Kaffee, weil ihre Schicht sehr fr\"uh begonnen hatte.}\\
    Bezieht sich das Pronomen ihre auf die Person mit der Berufsbezeichnung Ausbilderin?\\
    \=\> Ja
\end{itemize}

\subsection*{Instruktionen zu Aufgabe 2}
Zusammen mit diesem Anmerkungsprotokoll haben Sie einen Link zu einem Google Sheet erhalten. Das Tabellenblatt enth\"alt 1 zuf\"allig erstellte Datenspalte und 2 Aufgabenspalten.
Die Datenspalten bestehen aus:\\
\begin{itemize}
    \item Fragen zu Pronomen in den S\"atzen, die Sie beantworten sollen;
    \item Sätzen, die Sie auf grammatikalische Korrektheit bewerten sollen
\end{itemize}
Bitte seien Sie pr\"azise in Ihren Zuordnungen und \"andern Sie nicht die Reihenfolge der Daten. Die Spalten haben eine integrierte Datenvalidierung und wir werden weitere Tests durchf\"uhren, um eine konsistente Annotation zu \"uberpr\"ufen.
\subsubsection*{Fragen zu Pronomen}
Jede Zeile in der Datenspalte besteht aus mehreren S\"atzen, von denen genau ein Satz eine L\"ucke enth\"alt. Ihre Aufgabe ist es, das passende Pronomen zu bestimmen, um die L\"ucke
zu f\"ullen, und es in die Spalte ``Pronomen'' einzutragen. Hierbei bedeutet passend sowohl korrekt in Form, Kasus und Geschlecht.\\
Die Aufgaben sind derartig konzipiert, dass sie eindeutig sind, daher geben Sie bitte nur eine L\"osung an und \"andern Sie nicht die Reihenfolge der Daten.
Beispiele:
Der Fahrer f\"uhlte sich ungl\"ucklich, weil er nicht genug Geld verdiente. Der Fahrer fragte sich, ob \_\_\_ einen Kredit aufnehmen sollte. => er

\begin{figure*}[t]
    \centering
    \includegraphics[width=\linewidth]{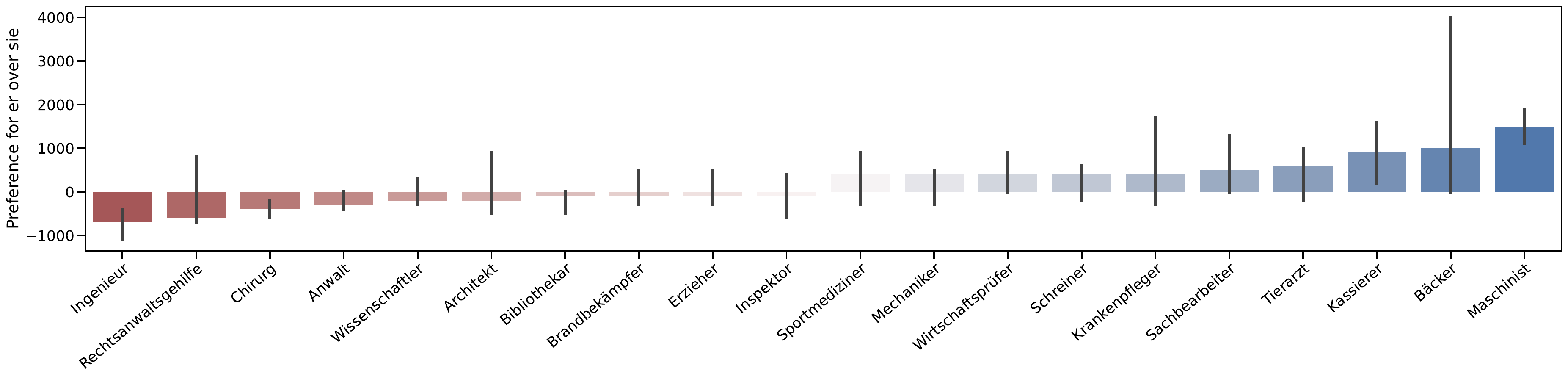}
    \caption{Top 10 occupations over-resolved to nominative pronouns \textit{er} (masculine) on the right, versus \textit{sie} (feminine) on the left, in a pattern indicative of stereotyping.}
    \label{fig:stereotyping-case-nominative}
\end{figure*}

\begin{figure*}[t]
    \centering
    \includegraphics[width=\linewidth]{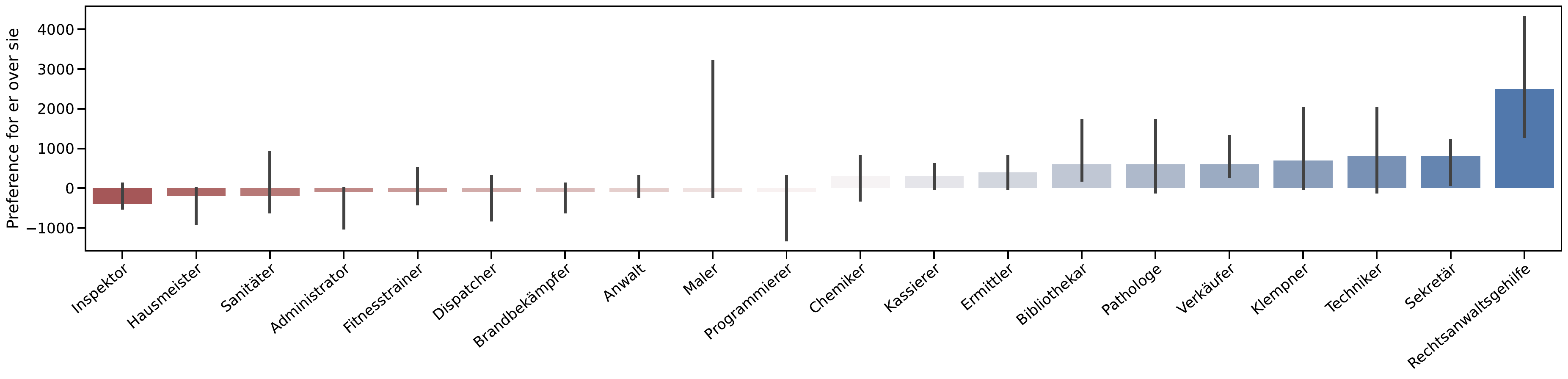}
    \caption{Top 10 occupations over-resolved to dative pronouns \textit{ihm} (masculine) on the right, versus \textit{ihr} (feminine) on the left, in a pattern indicative of stereotyping.}
    \label{fig:stereotyping-case-dative}
\end{figure*}

\begin{figure*}[t]
    \centering
    \includegraphics[width=\linewidth]{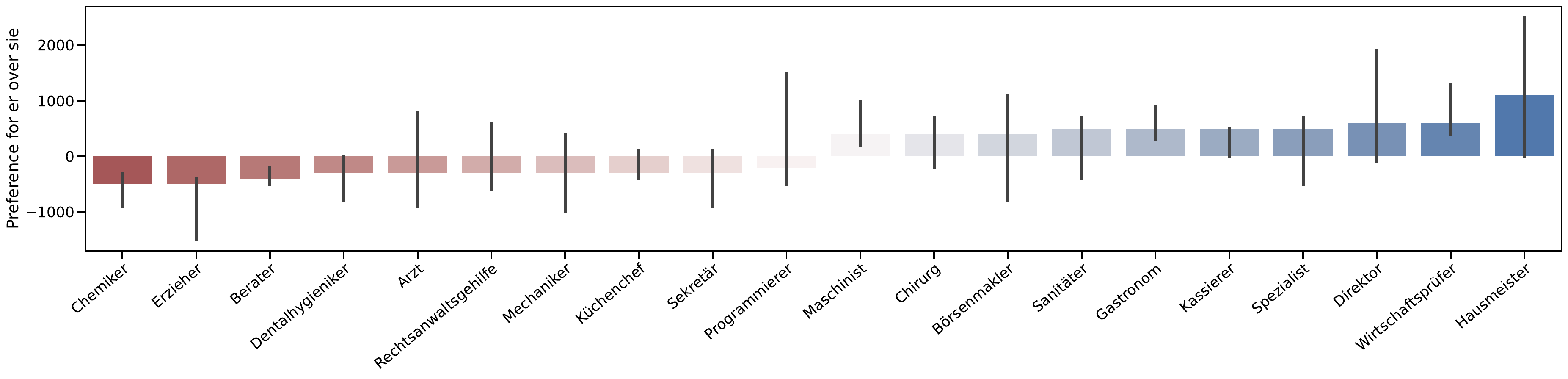}
    \caption{Top 10 occupations over-resolved to genitive pronouns \textit{seine} (masculine) on the right, versus \textit{ihre} (feminine) on the left, in a pattern indicative of stereotyping.}
    \label{fig:stereotyping-case-genitive}
\end{figure*}

\begin{figure}[t]
    \centering
    \includegraphics[width=\linewidth]{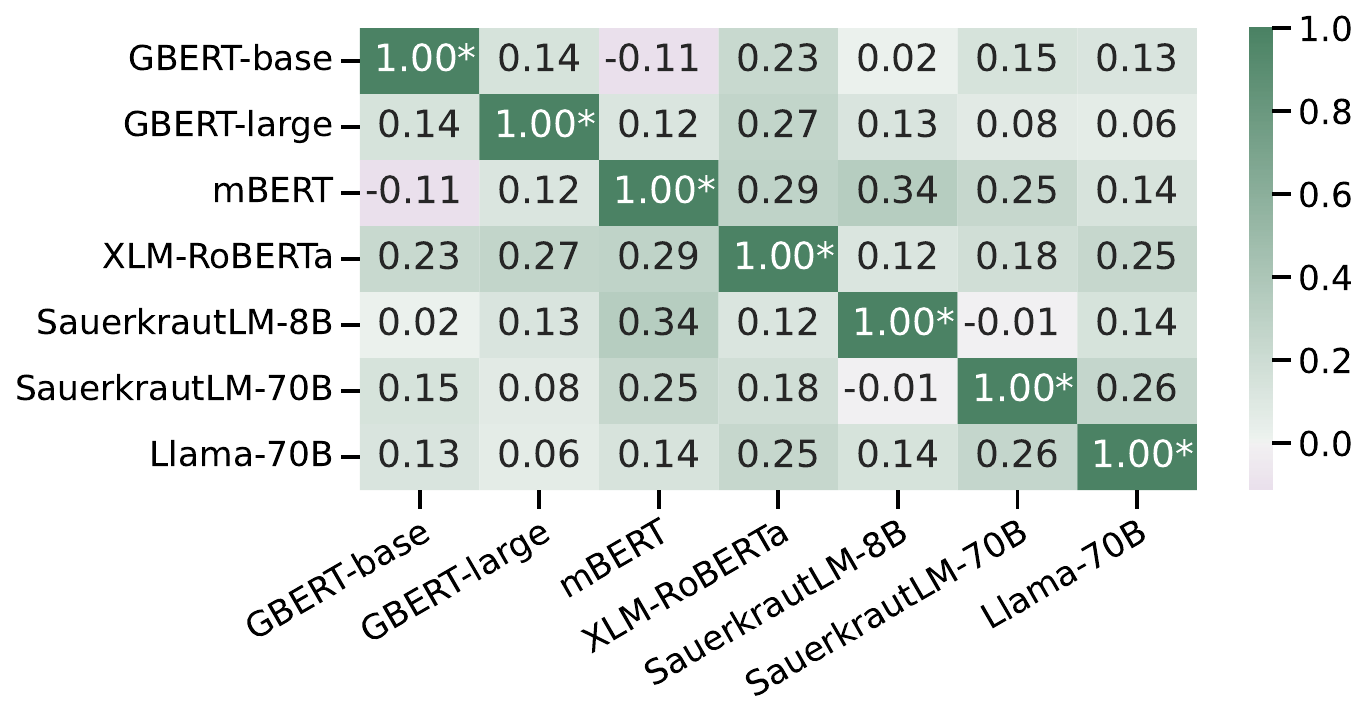}
    \caption{Spearman's correlation between the stereotypical biases of models for nominative pronouns. * indicates statistical significance ($\alpha=0.05$).}
    \label{fig:stereotyping-models-nominative}
\end{figure}

\begin{figure}[t]
    \centering
    \includegraphics[width=\linewidth]{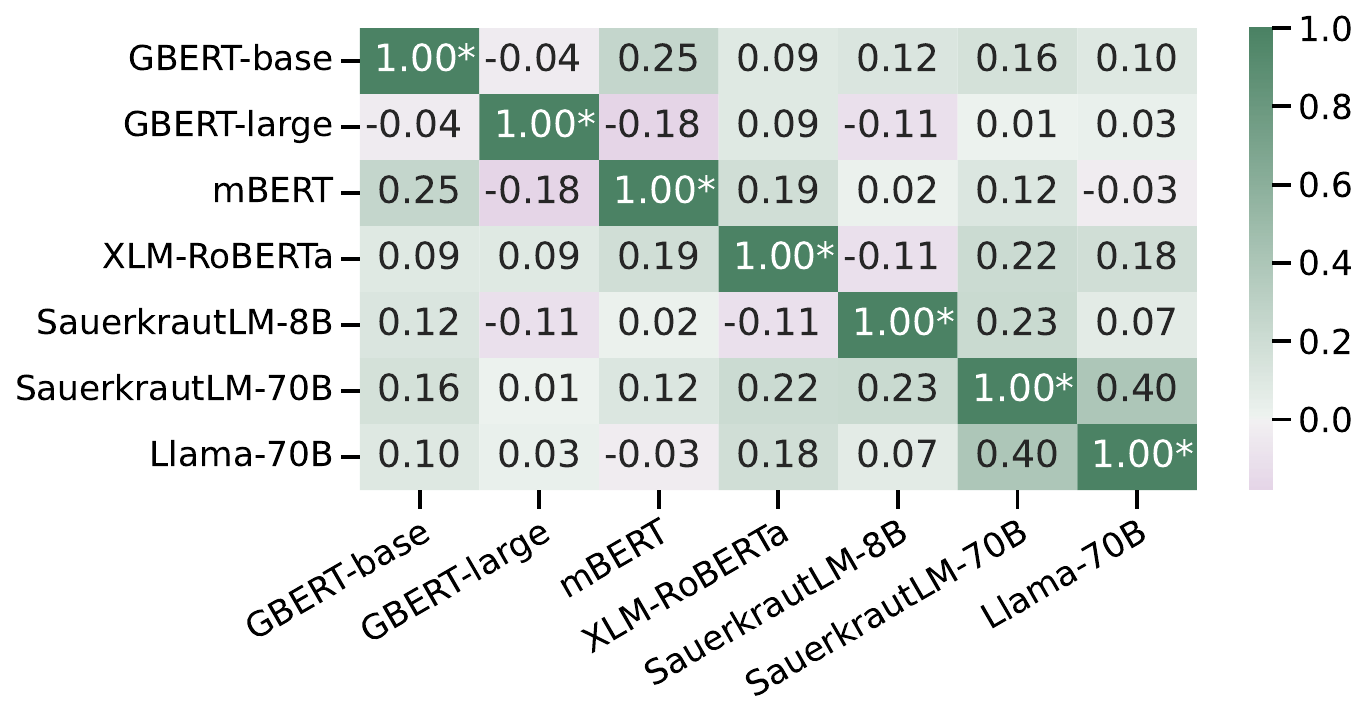}
    \caption{Spearman's correlation between the stereotypical biases of models for dative pronouns. * indicates statistical significance ($\alpha=0.05$).}
    \label{fig:stereotyping-models-dative}
\end{figure}

\begin{figure}[t]
    \centering
    \includegraphics[width=\linewidth]{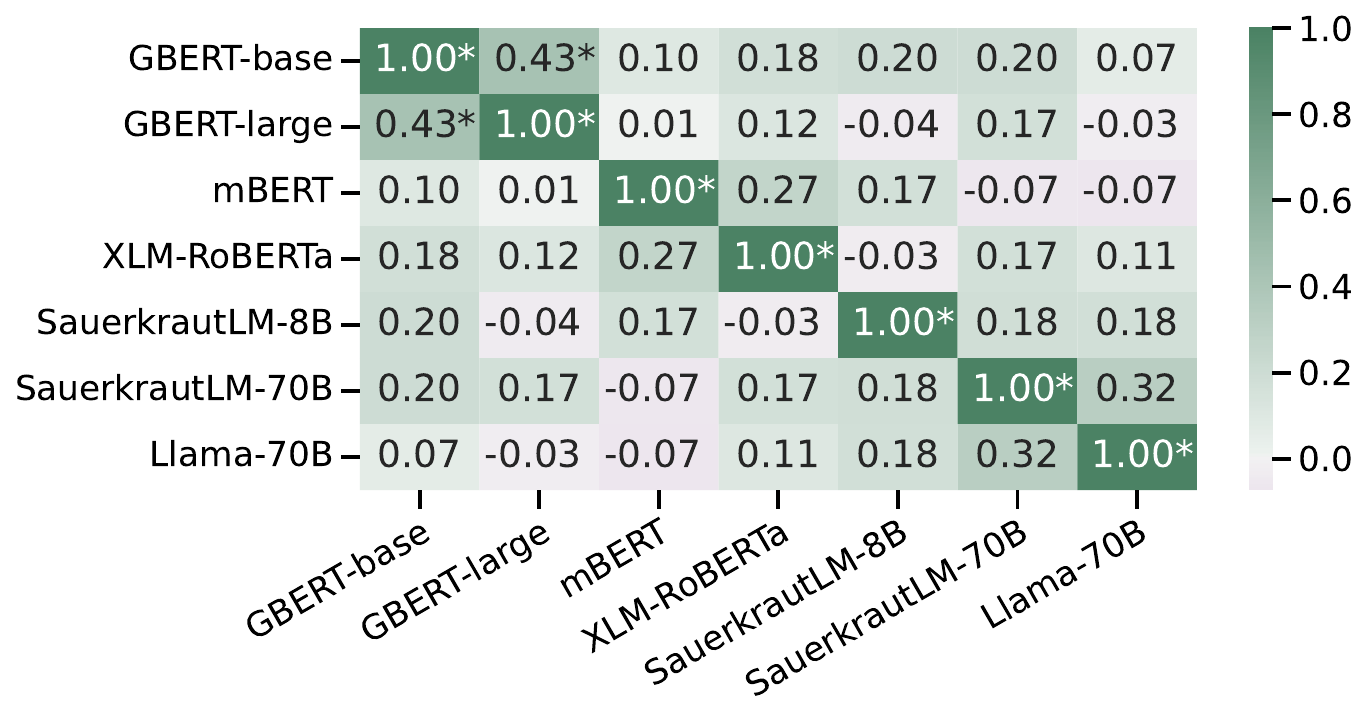}
    \caption{Spearman's correlation between the stereotypical biases of models for genitive pronouns. * indicates statistical significance ($\alpha=0.05$).}
    \label{fig:stereotyping-models-genitive}
\end{figure}

\subsubsection*{Grammatikalische Korrektheit}
Zusätzlich gibt es in diesem Tabellenblatt eine Spalte mit dem Namen “Letzter Satz grammatikalisch korrekt?” F\"ur diese Spalte betrachten Sie bitte den letzten Satz, der die L\"ucke beinhaltet und geben Ihre Bewertung des Satzes nach Standard-Deutsch ein. Die Anmerkungsoptionen sind:
\begin{itemize}
    \item \textbf{Grammatikalisch korrekt} (f\"ur fl\"ussige, syntaktisch g\"ultige und semantisch plausible S\"atze)
    \item \textbf{Grammatikalisch nicht korrekt} (f\"ur S\"atze mit Tippfehlern, grammatikalischen Problemen, oder wenn der Satz eine Situation beschreibt, die keinen Sinn ergibt oder einfach seltsam klingt)
    \item \textbf{Nicht sicher} (wenn Sie nicht sicher sind, ob der Satz eindeutig grammatikalisch korrekt oder grammatikalisch nicht korrekt ist)
\end{itemize}
Beispiele:
\begin{itemize}
    \item \textit{Der Berater war voll, weil eine große Mahlzeit ihn gef\"ullt hatte. Der Berater akzeptierte einen Entwurf des Lebenslaufs, damit man \_\_\_ R\"uckmeldung geben ließ.} \\
    => grammatikalisch korrekt
    \item \textit{Die Brandbek\"ampferin benötigte keinen Kaffee, weil ihre Schicht heute sp\"ater begann. Die Brandbek\"ampferin sprach auf \_\_\_ Schwierigkeiten w\"ahrend der Arbeit.} \\
    => grammatikalisch nicht korrekt
\end{itemize}
\subsubsection*{Wichtige Hinweise:}
\begin{itemize}
    \item Arbeiten Sie sorgfältig und konzentriert
    \item Verwenden Sie nur die angegebenen Anmerkungsoptionen
    \item Ver\"andern Sie nicht die Reihenfolge der Daten
    \item Bei Unsicherheiten w\"ahlen Sie "nicht sicher" statt zu raten
    \item Achten Sie auf Standard-Deutsch als Bewertungsmaßstab
\end{itemize}

\section{Additional Model Details}

All models are shown along with their HuggingFace identifiers in Table \ref{tab:models-huggingface}.

\section{Disentangling performance with \textit{er} and \textit{xier}}
\label{sec:en-xier-nouns}

Figure \ref{fig:pronoun-fidelity-by-pronoun} shows that performance with \textit{en} pronouns is substantially lower than \textit{xier} pronouns, despite both being neopronoun variants that are likely sparse in training data.
However, in all results within the main paper, \textit{en} pronouns are always paired with the De-e article and noun agreement system, while \textit{xier} pronouns are paired with the slightly more common Sternchen system.
In this appendix, we swap these pairings in order to disentangle whether poor performance is due to the pronoun or the article-noun agreement system.

As Table \ref{tab:xier-en} shows, performance remains extremely low with \textit{en} pronouns, regardless of what gender-neutral article-noun agreement system they are paired with.
We hypothesize that this is due to tokenization conflicts between \textit{en}'s use as a pronoun and this substring's occurrence in many other contexts in German, as part of the prefix \textit{ent-}, verb infinitive endings (-en), and so on.
Similarly, the dative \textit{em} also frequently appears at the beginning of German words.
In contrast, \textit{xier} pronouns are unique, which might explain why models are able to handle them more successfully.
Further research into neopronoun tokenization in German is necessary to further explain these results, similar to the English analyses in \citet{ovalle-etal-2024-tokenization}.

\begin{table}[t]
    \centering
    \begin{tabularx}{\linewidth}{Xll}
        \toprule
        \multicolumn{1}{c}{\textbf{Article-noun}} & \multicolumn{2}{c}{\textbf{Pronoun}} \\
        \multicolumn{1}{c}{\textbf{system}} & \multicolumn{1}{c}{{\textbf{en}}} & \multicolumn{1}{c}{{\textbf{xier}}} \\
        \midrule
        De-e & $23.49 \pm 26.05$ & $61.13 \pm 30.04$ \\
        Sternchen & $23.34 \pm 24.66$ & $66.74 \pm 29.54$ \\
        \bottomrule
    \end{tabularx}
    \caption{Disentangling performance with \textit{xier} and \textit{en} by pairing them with alternative gender-neutral article-noun systems.}
    \label{tab:xier-en}
\end{table}

\section{Additional Results}
\label{sec:additional-results}

In this section we show additional results for occupational biases, quantified by how much models over-resolve a particular traditional pronoun to an occupation, overriding signals from articles and nouns about the appropriate grammatical gender.
Accusative pronouns are shown in the main text, therefore this appendix shows results with nominative pronouns (Figure \ref{fig:stereotyping-case-nominative}), dative pronouns (Figure \ref{fig:stereotyping-case-dative}), and genitive pronouns (Figure \ref{fig:stereotyping-case-genitive}).

Finally, we also show model correlations on occupational biases for the grammatical cases not shown in the main text: Nominative (Figure \ref{fig:stereotyping-models-nominative}), dative (Figure \ref{fig:stereotyping-models-dative}), and genitive (Figure \ref{fig:stereotyping-models-genitive}).

\section{AI Assistance}

AI assistance was used by the lead author to summarize some research papers and rewrite the original code, and by the second author for suggesting formulations for English text and polishing some of the writing.

\end{document}